# Machine Learning-Based Detection and Analysis of Suspicious Activities in Bitcoin Wallet Transactions in the USA

Md Zahidul Islam[1], Md Shahidul Islam[2], Biswajit Chandra das[3], Syed Ali Reza[4], Proshanta Kumar Bhowmik[5], Kanchon Kumar Bishnu[6], Md Shafiqur Rahman[7], Redoyan Chowdhury[8], Laxmi Pant[9]

**Abstract**

*The dramatic adoption of Bitcoin and other cryptocurrencies in the USA has revolutionized the financial landscape and provided unprecedented investment and transaction efficiency opportunities. The prime objective of this research project is to develop machine learning algorithms capable of effectively identifying and tracking suspicious activity in Bitcoin wallet transactions. With high-tech analysis, the study aims to create a model with a feature for identifying trends and outliers that can expose illicit activity. The current study specifically focuses on Bitcoin transaction information in America, with a strong emphasis placed on the importance of knowing about the immediate environment in and through which such transactions pass through. The dataset is composed of in-depth Bitcoin wallet transactional information, including important factors such as transaction values, timestamps, network flows, and addresses for wallets. All entries in the dataset expose information about financial transactions between wallets, including received and sent transactions, and such information is significant for analysis and trends that can represent suspicious activity. This study deployed three accredited algorithms, most notably, Logistic Regression, Random Forest, and Support Vector Machines. In retrospect, Random Forest emerged as the best model with the highest F1 Score, showcasing its ability to handle non-linear relationships in the data. Insights revealed significant patterns in wallet activity, such as the correlation between unredeemed transactions and final balances. The application of machine algorithms in tracking cryptocurrencies is a tool for creating transparent and secure U.S. markets. As virtual currencies gain increased acceptance and transactions become increasingly sophisticated, machine algorithms can provide processing capabilities for enhancing supervision and compliance operations. Complicated algorithms can be programmed to search through massive sets of transactional information, identifying trends that could be indicative of fraud and compliance failures. With the use of past data, such algorithms can become trained to detect abnormalities in real-time, and regulators and financial institutions can respond promptly to suspicious activity.*

## Introduction

Islam et al. (2024b), reported that the increased use of Bitcoin and other cryptocurrencies over the past decade reoriented citizens' and companies' perceptions and use of money. As a virtual, decentralized form of money, Bitcoin grew into a sophisticated asset class, with a range of investors and users in America. Reduced transaction fees, rapid processing, and bypassing of traditional banking networks have aided in its widespread acceptance. Khan et al. (2024), found that Millions of Americans in 2023 use Bitcoin actively, with a range of wallets and exchanges to buy, sell, and exchange cryptocurrencies. All of this heightened demand for Bitcoin use and acceptance has fueled a sharp increase in transactions processed through its network. But precisely those same factors that make Bitcoin attractive—its anonymity and lack of

---

[1] MBA in Business Analytics, Gannon University, Erie, PA, Email: islam013@gannon.edu, (Corresponding Author)
[2] MBA- Business Analytics, International American University
[3] BS in Computer Science, Los Angeles City College.
[4] Department of Data Analytics, University of the Potomac (UOTP), Washington, USA.
[5] Department of Business Analytics, Trine University, Angola, IN, USA
[6] MS in Computer Science, California State University Los Angeles
[7] MBA in Management Information System, International American University
[8] MBA in Management Information System, International American University
[9] MBA Business Analytics, Gannon University, Erie, PA

centralization—generate tremendous headaches for financial regulators and law-enforcement agencies concerned with tracking and controlling the use of virtual currencies.

Rahman et al. (2024a), indicated that unlike traditional banking, whose transactions can, in many instances, be monitored and traced, Bitcoin transactions have a kind of pseudonymity: even when information about them is publicly posted in a blockchain, owners' identities go un-disclosed. As a consequence, such lack of transparency generates a rich field for illicit activity, including fraud, evasion of taxes, and money laundering. As many studies have concluded, a considerable proportion of Bitcoin transactions have an illicit origin, and regulators and security professionals have run alarm bells over them. The methods with which criminals exploit cryptocurrencies have kept pace with them, and conventional techniques, with a strong bias towards techniques in terms of compliance with rules and observation in a traditional manner, have become less effective in countering Bitcoin transactions' complexity and volumes. Conventional techniques fail to keep pace with increasingly sophisticated techniques embraced by cybercrimes, and many suspicious activity, therefore, goes undetected. All of this throws a spotlight onto a critical imperative for new techniques that can effectively detect and monitor potential peril in Bitcoin transactional ecologies (Sizan, 2023b).

*Problem Statement*

According to Shawon (2024), the growing use of Bitcoin wallets for a range of transactions, including illicit ones, creates a key challenge for regulators and America's law-enforcement agencies. There have been indications that a significant portion of Bitcoin is processed through wallets with a nexus for illicit use, and its use in illicit financing, terrorism, and other frauds is a concern. Concerning this, it is important to have effective tools for tracking such suspicious activity in a way that will secure financial structures and preserve compliance with existing regimes of regulation. As per Jui et al. (2023), the drawback of traditional detection techniques can be understood in their inadaptability to changing times concerning Bitcoin transactions. Rule-based techniques, in most instances, rely on predefined suspicious behavior signatures and can become rigid and ineffective in dealing with sophisticated and flexible techniques embraced by offenders. Besides, Bitcoin network transactions can become voluminous, and thus, impractical for observation in a manual manner. Consequently, an imperative for complex analysis techniques capable of leveraging the potential of machine learning in supporting suspicious activity detection in Bitcoin wallet transactions arises.

*Research Aim*

The prime objective of this research project is to develop machine learning algorithms capable of effectively identifying and tracking suspicious activity in Bitcoin wallet transactions. With high-tech analysis, the study aims to develop a model with a feature for identifying trends and outliers that can expose illicit activity. Developing such algorithms will not only simplify real-time suspicious transaction identification but will produce actionable intelligence capable of driving financial security and compliance with regulators in America. To serve such an objective, the study will explore a variety of algorithms in machine learning and implement them in Bitcoin transaction data in a try to comprehend its challenge. By analyzing past behavior in transactions and identifying key factors present in suspicious activity, the work will seek to build a comprehensive model with an ability to adapt to cryptocurrencies' ever-changing nature of transactions. In conclusion, this work will seek to contribute towards a broader dialogue of financial security and compliance in Bitcoin and cryptocurrencies in general.

*Scope and Relevance*

The current study specifically focuses on Bitcoin transaction information in America, with a strong emphasis placed on the importance of knowing about the immediate environment in and through which such transactions pass through. The application of machine algorithms for suspicious activity-related risk identification and evaluation is particularly relevant in consideration of increased regulatory concern over cryptocurrencies. As regulators try to introduce a balanced regime that encourages innovation but at the same time curtails danger, information derived through such a study can serve as a useful tool for regulators, lawmen, and financial entities. The impact of this work extends past its direct Bitcoin transactional sphere. As cryptocurrencies mature and become increasingly prevalent, methodologies developed in this work can have a potential impact on future cryptocurrencies and general use in blockchain technology. By offering a baseline for an awareness of how machine learning can be leveraged for suspicious activity, this work aids in contributing to current work in securing and fortifying the financial system overall.

## Literature Review

*Bitcoin and Cryptocurrency Adoption in America*

Abdirahman et al. (2024), reported that the use of Bitcoin and cryptocurrencies in America has experienced a tremendous journey ever since Bitcoin began in 2009. Bitcoin initially began as a niche, with technology enthusiasts and early adaptors being its largest followers initially. As awareness and familiarity with cryptocurrencies increased, acceptance in most industries, including retail, finance, and investments, received a significant boost. As of 2023, millions of Americans utilize Bitcoin, utilizing it for payments, investments, and stores of value, a sign of a larger migration toward digital assets. Bitcoin volumes in transactions have gone through a tremendous boost, with volumes in a single day in hundred thousand and even a billion dollars in value in most cases (Al Montaser et al., 2025) This high rise in volumes in transactions is a sign not only of the increased use of Bitcoin but even its viability as a financial tool for the general public.

In contrast to such expansion, the U.S. regulatory environment for cryptocurrencies is complex and disorganized. There are a variety of state and federal agencies involved in regulating activity in cryptocurrencies, and each one of them comes with its compliance requirements and sets of laws. All these quilted laws make it a challenging environment for regulators and even for users when it comes to tracking and compliance with cryptocurrencies ( Nayyer et al., 2023). One of the largest obstacles is that cryptocurrencies have a decentralized nature, and in such a case, no single entity, including a bank, is involved in a transaction. As much as such decentralization is a defining feature for attracting users, it proves to make it difficult for regulators to implement laws and monitor transactions effectively.

Moreover, technological advancement in cryptocurrencies occurs at a pace that tends to outdo frameworks for governance in terms of reaction, and a governance lag then sets in, with new technology and use at a pace outdoing new governance structures and controls in position. Cryptocurrencies are anonymous, and with anonymity, a challenge for regulators in tracking illicit activity, even with an accessible public record in a blockchain, with owners of a wallet not being identifiable, and therefore, not trackable, for illicit activity tracking. All these necessitate new and alternative methodologies in terms of governance and tracking, and therefore, creating strong tools for tracking suspicious activity in Bitcoin transactions is imperative (Nerurkar, 2023).

*Suspicious activity in Bitcoin wallets*

Pocher et al. (2023), contended that the success of Bitcoin, unfortunately, has been accompanied by an increased use for illicit purposes through its use. Cryptocurrencies have become a focal point for illicit use for a variety of illicit purposes, including fraud, illicit trading, and money laundering. Money laundering, in particular, has become a concern, with criminals exploiting Bitcoin's pseudo-anonymity in an attempt to cover illicit funding sources. Estimates have seen a high level of Bitcoin transactions concerning illicit use, with a cry for increased monitoring and enforcement having been heard in response.

Rahman et al. (2025), postulated that fraud operations in the environment of cryptocurrencies run in abundance, in many forms such as Ponzi operations, scams through phishing, and pump-and-dump operations. There is no regulating mechanism, and with cryptocurrencies' high technicality, fraudsters, and scammers have a fertile breeding ground for exploiting innocent investors. Illegal trading operations such as washing and trading through insiders make it even more complex, undermining the integrity of trading platforms for cryptocurrencies.

The establishment of suspicious activity in cryptocurrencies entails the identification of specific markers that can potentially refer to illicit activity. Common markers for suspicious activity include rapid fund transfers between addresses, activity with addresses with a documented record of illicit use, and unorthodox trading behavior not in compliance with traditional trends (Rahouti et al., 2018) For instance, large transactions that occur in a condensed timeframe and have no perceivable economic purpose can raise concern, and, as such, a deeper investigation is merited. Interaction with mixing platforms, whose purpose is to make a record of transactions untraceable, is a prevalent sign of illicit activity and, as such, deserves heightened scrutiny.

The difficulty in tracking such suspicious activity is compounded by the high level of transactions taking place in the Bitcoin network, with thousands processed in a single minute alone. Monitoring each one individually is becoming increasingly impractical, and it is in such a scenario that one realizes the necessity for automated tools that can scan through transaction information in bulk and detect any danger and abnormalities present in them (Padgorelec et al., 2018).

*Machine Learning in Fraud Detection*

Machine learning is a powerful tool in anti-fraud operations, offering new methodologies for discovering suspicious behavior and outliers in a variety of industries, including cryptocurrencies, through its big datasets and complex algorithms. With complex algorithms and big datasets, fraud detection capabilities can detect trends that can represent fraud, and enormously enhance fraud capabilities (Shahbazi & Byun, 2022).

Applications of machine learning for fraud detection include a variety of methodologies, including supervised, unsupervised, and reinforcement learning. Labeled datasets can be trained in a model in supervised learning, and then, through such training, can learn about real and fraudulent transactions. With such training, predictive models can classify new transactions in terms of discovered patterns (Saeidimanesh, 2024). In contrast, unsupervised methodologies in machine learning can serve to discover outliers in datasets with little availability of examples with labels. Unsourced examples in such datasets can be discovered through unsupervised methodologies, and such examples can represent outliers in transaction datasets, and hence, can represent fraud examples, and such examples can then deserve investigation.

Sizan et al. (2025), argued other types of machine algorithms have been implemented in the analysis of cryptocurrency transactions in a quest to enhance capabilities for detection. Decision trees, neural networks, and support vector machines have been effective in identifying suspicious activity. For instance, deep neural networks, with the ability to analyze complex trends in big data, have been implemented in fraud detection in transactions through analysis of behavior and activity. How machine algorithms can update and become even more effective throughout ongoing training is an added improvement in terms of new and emerging threats discovery. Whereas its application in fraud detection is not encumbered with several impediments, its application in cryptocurrencies is hindered by several impediments.

Since cryptocurrencies' value is calculated in a dynamic environment, and criminals' methodologies change with them, updating and retraining machine algorithms is a continuous necessity. Machine algorithms' interpretability can become a problem for compliance with regulators, whose justification requirements for actions taken through such algorithms can at times become a problem for them. How one can have effective detection capabilities and comply with regulators at the same time is a critical consideration in creating such a solution through machine algorithms (Nerurkar, 2023).

*Research Gaps*

Islam et al. (2024b), asserted that although a significant amount of work in fraud detection with machine learning has been performed, gaps in such work, particularly in the U.S.-based Bitcoin fraud detection case, have not yet been filled in. Most fraud detection work with machine learning to date has been in theoretical frameworks or general scenarios and not in specifically dealing with the U.S.-based challenge posed by its financial and state-by-state laws and legislation. With its state-by-state laws and complex financial system, the U.S. environment poses a challenge in terms of suspicious activity in Bitcoin transactions, and therefore, specific methodologies must be developed for its case. Thus, a strong demand for empirical work in studying the use of machine learning in U.S.-based Bitcoin fraud detection arises.

Moreover, most studies lack a synthesized perspective that integrates detection, risk analysis, and policy analysis. Jui et al. (2023), held that successful fraud detection is not a case of searching for suspicious transactions but, in reality, involves an awareness of its larger implications for financial security and compliance with laws and regulations. There is a need for studies researching in detail how algorithms for machine learning can integrate with present compliance and regulatory frameworks in a manner that fortifies compliance and monitoring processes. Bridging the gap between capabilities for detection and policy analysis will allow researchers to gain a deeper perspective of how Bitcoin and similar cryptocurrencies can have their respective risks lowered.

*Data Collection and Exploration*

The dataset is composed of in-depth Bitcoin wallet transactional information, including important factors such as transaction values, timestamps, network flows, and addresses for wallets. All entries in the dataset expose information about financial transactions between wallets, including received and sent transactions, and such information is significant for analysis and trends that can represent suspicious activity. Public blockchain datasets make up the principal sources of information, offering unchangeable information about all Bitcoin transactions, and information extracted through extraction of information through cryptocurrency exchange APIs, offering supplementary information about the behavior and trading activity of users. Besides, information extracted through the extraction of information through blockchain analysis platforms enriches the dataset, offering enriched information, including types of wallets and risk assessments, and offering a deeper analysis of trading flows and illicit activity in Bitcoin networks.

*Data Pre-Processing*

The code snippet portrayed several routine operations in preprocessing data. First, it addressed incompleteness through an analysis of missing values in each column and then dropped rows with a missing 'address' or 'hash160' field. Secondly, it identified and dropped duplicates in case of any, in an attempt to make the data unique. Thirdly, it computed the 'address' and 'hash160' column types to string, assuming them to serve as identifiers, and thus normalize them for future analysis and modeling work. All these operations in concert attempted to cleanse and make the data ready for use by dealing with incompleteness, redundancy, and type incompatibility.

*Exploratory Data Analysis (EDA)*

Exploratory Data Analysis (EDA) is a critical initial stage in the research process, including describing, visualization, and interpretation of a key feature of a dataset. EDA employed a variety of statistical and graphical tools for discovering relations, outliers, and patterns in the data, in a position to enable deeper insights into the underlying phenomena, even before sophisticated statistical techniques and model approaches can be utilized. EDA helped in discovering potential hypotheses, key variables, and research questions, guiding future studies and experiments' planning. Besides, EDA helped in discovering errors, contradictions, and gaps in the data, in a position to enable proper preprocessing and cleaning operations. Moreover, EDA helped in discovering hidden patterns, trends, and outliers, not necessarily uncovered through traditional statistics, and new information and breakthroughs followed.

*Correlation Heatmap of Numerical Features*

The Python script generated a numerical feature correlation heatmap for a Pandas Data Frame df. First, matplotlib.pyplot and seaborn for plotting are loaded. Next, a list of numerical columns was declared, specifying numerical feature column(s) to use for analysis. Subsequently, a correlation matrix for these numerical feature column(s) with df[numerical_cols].corr() was generated. Afterward, a heatmap for these numerical feature column(s) correlations was generated with seaborn.heatmap(), with annot=True plotting values in cells of the heatmap, format='.2f' specifying two-digit value format in cells, cmap='coolwarm' specifying cool and warm colors, and cbar=True plotting a color bar for visualization. The generated heatmap, a graphical representation of the correlation matrix, was then displayed with plt.show() following a title with plt.title(). With this visualization, the researcher obtained a glance to identify high-correlated feature(s), beneficial for feature selection and model development in future analysis.

*Output*

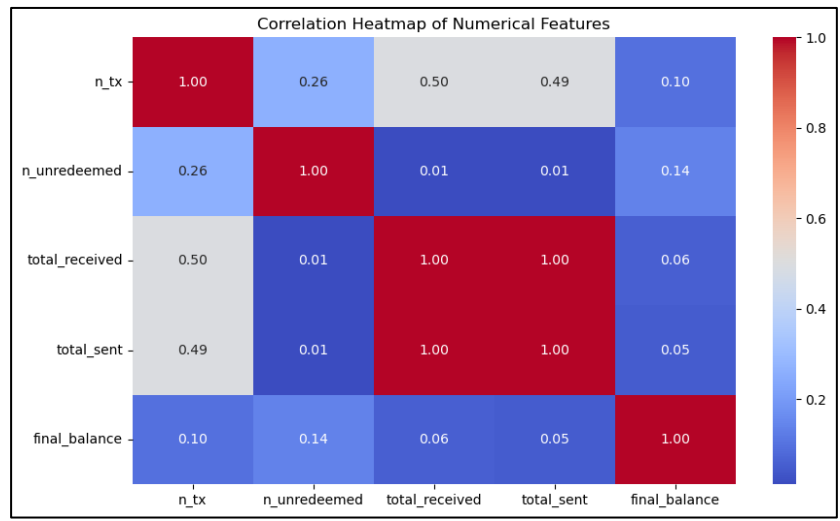

**Figure 1. Portrays Heatmap of Numerical Features**

The correlation heatmap reveals strong relations between a variety of numerical factors in Bitcoin wallet transactions. In particular, "total_received" is positively correlated with "n_tx" (0.50) and "total_sent" (0.49), with a sign that wallets with a high number of transactions receive high values of Bitcoin. In contrast, "total_sent" is weakly correlated with "final_balance" (0.06), and, therefore, the overall value sent doesn't necessarily affect wallets' values in terms of residuals. "n_unredeemed" moderately corresponds with "n_tx" (0.26) and "total_sent" (0.14), with an indication of a relation between unredeemed and overall activity in wallets. Overall, the heatmap validates the interrelated nature of values in transactions, with a high value placed in received and sent totals in explaining behavior in wallets.

*Top 10 Wallets by Total Received*

The code script in Python identified and plotted the top 10 wallets in terms of received and sent values in a Data Frame df. First, it took the 10 largest values for 'total received' and 'total sent' with the largest () and stored them in top_received and top_sent, respectively. Second, it created a figure with two plots using matplotlib.pyplot with two subplots. The first plot in the left subplot was a horizontal bar plot for the top 10 wallets in terms of 'total received' using seaborn.barplot() with 'address' for y and 'total received' for x, and a reversed Blues palette. Likewise, the plot in the right subplot was a bar plot for the top 10 wallets in terms of 'total_sent' with a reversed Greens palette. Both plots had titles and axis labels for ease of reading, and proper spacing between plots is guaranteed with plt.tight_layout(). Ultimately, plt.show() plotted the overall figure, allowing for a comparative view of top wallets in terms of received and sent values.

*Output*

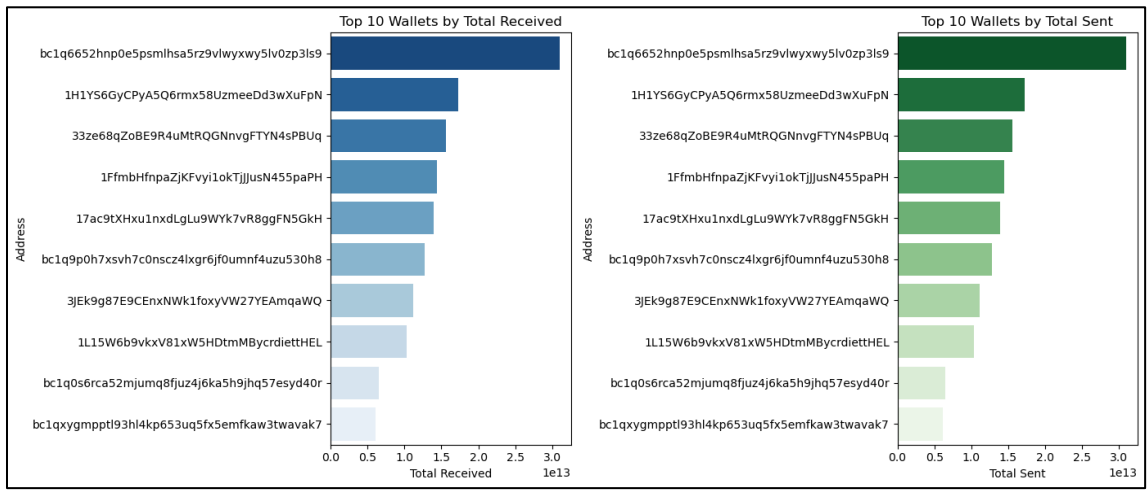

**Figure 2. Displays Top 10 Wallets by total Received**

The accompanying histogram presents two comparative horizontal bar plots of the top 10 wallets in terms of received (left) and sent (right) totals. Two sets of 10 wallets, in descending order, occur in both plots, with both plots consisting of the same 10 wallets in terms of respective volumes in terms of received and sent totals. In the "Top 10 Wallets by Total Received" plot, a dominant "bc1q6652hnp0e5psmlhsa5rz9vlwyxwy5lv0zp3ls9" with a much larger received total value stands out in comparison with other wallets. In contrast, a "1H1YS6GYCPYA5Q6rmx58UzmeeDd3wXuFpN" occupies the largest value in "Top 10 Wallets by Total Sent" with a relatively less sharp fall in values in terms of totals sent in the case of mentioned wallets. Interestingly enough, a variation in ordering between two plots proves that a high received total doesn't necessarily mean a high sent total, and a high sent total doesn't necessarily mean a high received total, respectively. Two plots have a dark-to-light color gradation, representing relatively larger values in terms of totals, even though numerical values for volumes in terms of received and sent totals don't occur in the accompanying picture of plots.

*Total Sent vs. Total Received*

The code script in Python generated a scatter plot to visualize 'total received' and 'total sent' values, with 'n_tx' being represented for point size and 'final balance' for colors in terms of a "viridis" colormap. It generated a figure with a specific size using plt.figure(). It constructed a visualization with a call to a function in seaborn, namely, a scatterplot (), mapping 'total_received' for x, 'total sent' for y, 'n_tx' for specifying point size, and 'final_balance' for specifying colors in terms of a "viridis" colormap. It sets a title, axis labels, and a legend placed out of the plot region for easier visualization using bbox_to_anchor. It then sets plotted parameters for a tight layout using plt.tight_layout(), and plots using plt.show(), providing insights into entities' transaction behavior and distribution of balance.

*Output*

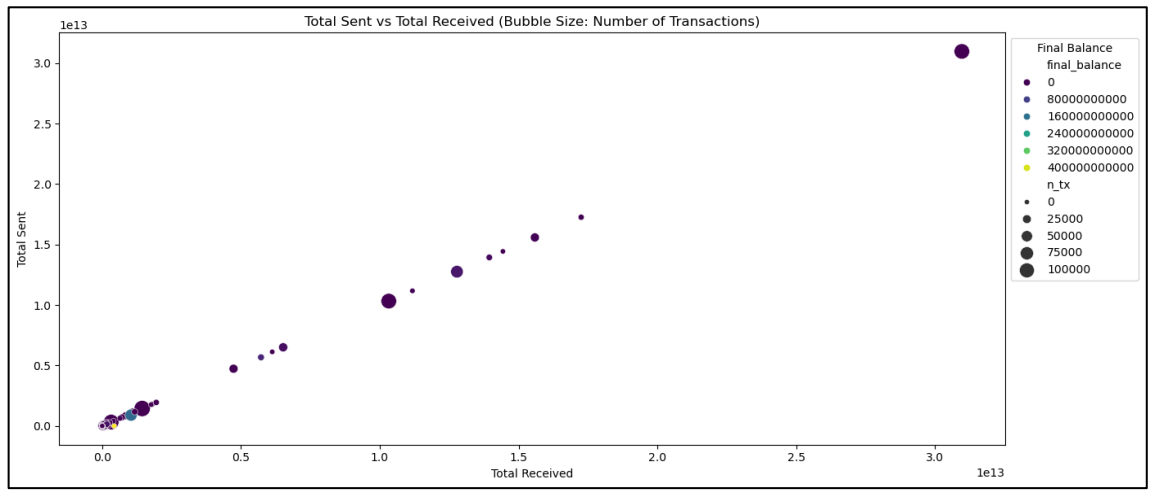

**Figure 3. Total Sent vs. Total Received**

The bubble plot shows received and sent totals in Bitcoin transactions, with each bubble representing respective numbers for each wallet. In general, a strong positive correlation can be discerned, with larger received totals in wallets tending towards larger sent totals, indicative of active use in Bitcoin's economy. Most striking, larger bubbles, representing wallets with long transactional histories, cluster towards the upper right, meaning not only receive high totals of Bitcoin but actively send high totals as well. Most striking, a few outliers can be noticed with incredibly high sent totals but relatively low received totals, representing high outflows, possibly trading activity, and possibly even withdrawals. Having a range of final totals between bubbles enriches analysis, providing a glimpse into liquidity and operational activity in these wallets in the marketplace.

*Distribution of Highly Skewed Columns*

The code snippet defined a function plot_log_distributions to visualize the distributions of highly skewed data columns using a logarithmic scale. The function took a data frame, a list of column names, and a title as input. Inside the function, it iterated through the specified columns, creating a subplot for each. For every column, it applied a logarithmic transformation using np.log1p() (which adds 1 to avoid log(0)) and then generated a histogram using seaborn. His plot () with KDE (Kernel Density Estimate) enabled and a purple color. Each subplot was titled with the log-scaled column name, and the overall figure was given a super title. Finally, it called this function with the DataFrame df and a list of skewed columns ('total_received', 'total_sent', 'final_balance') to generate and display the log-scaled distribution plots for these columns.

*Output*

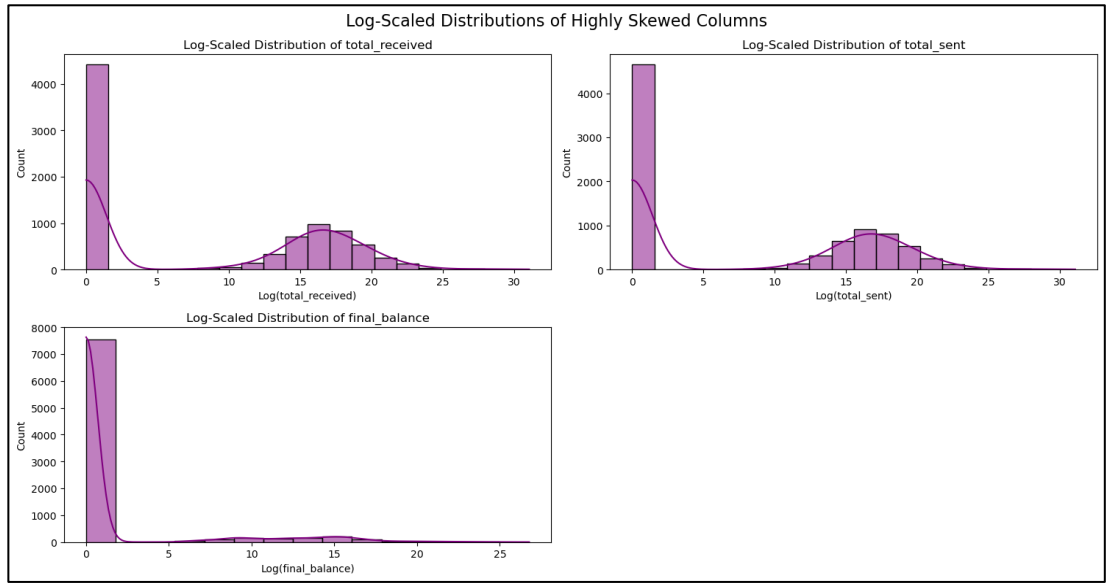

**Figure 4. Depicts Log-Scaled Distribution of Highly Skewed Columns**

The log-scaled distributions in the plot reveal extreme skews in Bitcoin transaction activity, namely for the variables total_received, total_sent, and final_balance. In the upper-left, a high peak near zero in the distribution for total_received reveals a high density of low-value reception values for many wallets, with a long tail towards larger values, representing a small group of wallets with a lot of Bitcoin received. Similarly, in the upper-right, a similar distribution for total_sent reveals a high density of relatively low-value transmissions, with most wallets sending little, a small group sending a lot, and a long tail towards larger values, representing a small group of wallets with a lot of Bitcoin sent out. In the lower region, a similar distribution for final_balance reveals similar skews, with a high density of many wallets with little and a small group with a lot, and a long tail towards larger values, representing a wealth inequality in Bitcoin owners. Overall, these log-scaled distributions reveal an unequal distribution of Bitcoin holdings and transmissions, with a high presence of a small group dominating the environment.

*Top 15 Wallets by Activity Intensity*

The provided code fragment calculated and plotted activity intensity for wallets in terms of the proportion of 'n_tx' to 'total received', and then ranked them in descending manner to choose the most active ones. First, it calculated activity intensity for each wallet with an added 1 in the denominator for excluding zero in case of 'total received', and then ranked them in descending manner. wank(ascending=False). It stored them in a new column 'activity intensity rank' in DataFrame df. Next, it plotted a horizontal bar plot with a plot of the 15 most active wallets with high activity intensity rank. With seaborn.barplot(), it plotted address' in y and its respective 'activity_intensity_rank' in x, with "rocket" colors for colors for bars. It is titled, labels, and then plotted with plt.show(), providing a concise view of most active wallets in terms of calculated intensity value.

*Output*

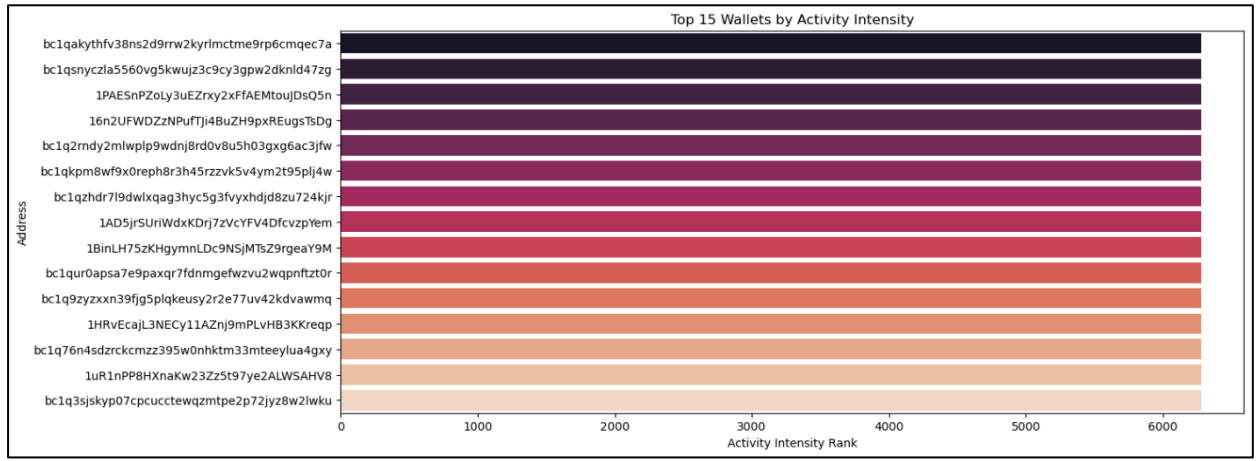

**Figure 5. Top 15 Wallets by Activity Intensity**

As showcased above, the 15 most active wallets ranked in terms of "Activity Intensity" are represented in the horizontal bar chart, a value derived both from transactions and received totals. "bc1qakythfv38ns2d9rrw2kyrlmctme9rp6cmqec7a" is seen with the largest activity intensity, with its bar extending out in front of everyone else, a sharp drop in activity intensity in between, and then a slow, steady drop in activity intensity for everyone else. "Activity Intensity Rank" appears to represent a calculated rank and not a raw activity intensity value, but with no labels, one can't say for sure that a larger rank corresponds to a larger intensity, but a comparative value seems to have been calculated for it. No calculation for this rank is shown, but a comparative activity value seems to have been calculated for it. Darker colors represent a larger activity intensity in an effectively used gradient, with lighter colors representing less activity intensity. There is a clear visualization of a hierarchy of activity in a simple, effective form, with a strong contrast between most and least active in this top 15 group.

*Number of Unredeemed Transactions and Final Balance*

The implemented Python script created a scatter plot with a linear regression line to plot the relation between 'n_unredeemed' and 'final_balance' in a Data Frame df. It created a figure with a specific size with plt.figure(). The heart of the plot is seaborn.regplot(), created a scatter plot of both variables fitting a linear model through them and plotting a regression line.'scatter_kws={'alpha':0.5}' controls the transparency of scatter points for easier visualization in case of overlaps, and 'line_kws={'color':'red'}' colors the regression line in red. The plot was then decorated with a title and axis labels for ease of visualization. Lastly, plt.show() plots and showed the generated plot, providing a basis for a visual evaluation of the relation and any linear relation between 'n_unredeemed' and 'final_balance'.

*Output*

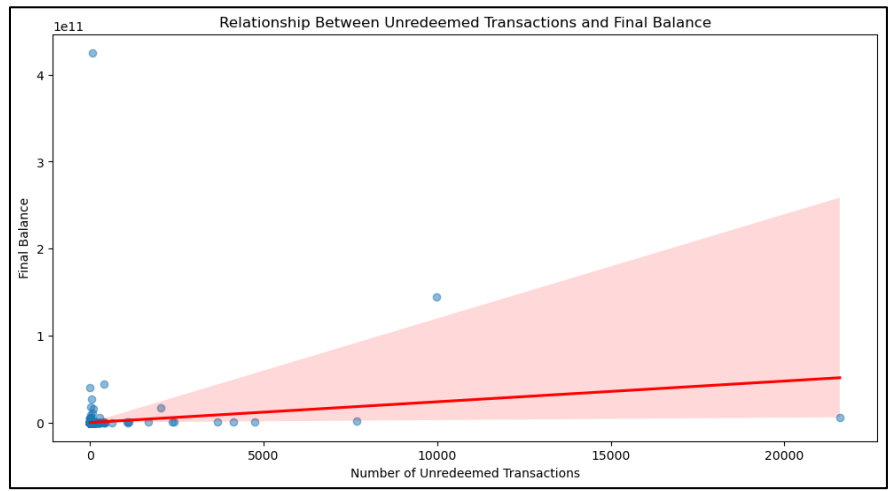

**Figure 6. Number of Unredeemed Transactions and Final Balance**

The above scatter plot reveals an emerging relation between the count of unredeemed transactions and the Bitcoin wallet's final value, with an emerging positive relation between them. As an increased count of unredeemed transactions, an increased direction in final values can be seen, with an emerging relation between them, with an indication that wallets with an increased count of unredeemed transactions have larger Bitcoin values in them. There is a positive relation between these two values supported through an emerging positive relation in the fitted regression line, drawn in red, with most values in both groups clustering at low values, with an indication that most wallets don't have an increased count of unredeemed transactions. There is an emerging filled region, with an emerging region of high activity, with an indication that wallets with an increased count of unredeemed transactions will have much larger values in them. Overall, the plot reveals an emerging impact of unredeemed transactions in terms of having an impact in terms of values in wallets, with emerging behavior in terms of holding high values of unredeemed funds in wallets.

## Methodology

*Feature Engineering*

Feature engineering is a critical activity in strengthening the predictive capabilities of machine learning algorithms, specifically in Bitcoin wallet analysis. One of the key features developed is transaction frequency, which describes the frequency at which a wallet transacts over some time. By defining the count of transactions per week, for instance, per week, per month, etc., we can differentiate between active and inactive wallets, and such differentiation is important in terms of describing the behavior of a user. Next, we inspected amount trends through an examination of transaction size distribution to detect trends in sending and receiving Bitcoin in terms of wallets' behavior. Important statistics such as mean, median, and variance of transaction values inform whether a wallet partakes in high-value deals or numerous, low-value deals. Next, we extract features representing a proportion of big deals and small deals, and such a proportion can reveal a variety of user behaviors—traders, long-term investors, etc. Not only do such engineered features enrich a dataset, but such engineered features can make a model even better at identifying outliers, and in fact, such engineered features can make a model even better at identifying trends and outlier's indicative of suspicious behavior, such as fraud, for instance.

We utilized conventional and graph-based features together, traditional ones for general information and graph-based ones for analyzing relations between wallets and transaction flows in the network. By projecting wallets onto nodes and transactions onto edges, we form a transaction graph, and through it, connectivity and interaction between wallets and between them can be examined, respectively. By utilizing, for example, a degree of centrality, a direct neighbor counts a wallet possessed, and important wallets in a network can be detected. To comprehend, moreover, a wallet's connectivity in terms of its distance to any other wallet and its role in acting in between transactions, closeness, and betweenness centrality, respectively, are calculated. By employing such sophisticated features, such hidden relations and transaction flows cannot necessarily be discovered with traditional feature analysis alone. By combining such sophisticated features with our dataset, a deeper picture of Bitcoin's ecosystem, with information about wallets' behavior towards each other and with a determination of key and suspicious nodes, can deserve a deeper investigation in a future analysis. With such a thorough feature engineering, a strong foundation for future modeling work is developed, and consequently, our analysis accuracy and solidity will increase.

*Overview of a Machine Learning Algorithm*

Random Forest is an algorithm that generates numerous decision trees during training and then spits out their prediction mode. It is a powerful algorithm for classification and overfitting resistance, and thus perfect for our dataset, possibly with a high feature count via our feature engineering. Random Forest can even handle numerical and categorical values and thus can have a mix of many features that have been engineered.

Logistic Regression is a simple model and one most commonly used for problem-solving in binary classification. Logistic Regression will not model complex relations with such a high level of accuracy as Random Forest but can interpret feature importance in a simple and computationally less expensive manner. With such a high-dimensional feature space, Logistic Regression can serve as a baseline model and can validate complex model output.

The Support Vector Machines (SVMs) are powerful classifiers with high performance in high-dimensional spaces. SVMs try to classify and obtain the best hyperplane that separates classes in feature space. SVMs can model non-linear relationships effectively with a kernel function, and such a model will best serve our analysis, specifically in its high accuracy in cases of binary classification, for which SVMs have been effective and powerful classifiers in practice.

*Training and Testing Framework*

We performed a robust training and testing environment in place to ensure integrity and reliability in our models. For this, individual training, testing, and validation sets for the dataset, and cross-validation techniques, are utilized. Initially, the dataset is divided into three parts: the training set, which is used to train the models; the validation set, which is utilized to tune model parameters and prevent overfitting; and the testing set, which serves as a final evaluation of model performance. A common practice was to allocate approximately 70% of the data for training, 15% for validation, and 15% for testing. This distribution allowed for a sufficient amount of data to train the models while reserving enough for rigorous evaluation. To make model robustness even more powerful, we apply cross-validation techniques, namely k-fold cross-validation. In k-fold cross-validation, training datasets were split into k folds, and k-1 folds were trained and validated one-fold at a time. Cross-validation was performed k times, with one-fold working as a one-time validation set in each case. All such individual values are averaged to obtain a sounder estimation of model performance. By using k-fold cross-validation, overfitting can be circumvented and a truer estimation of model performance in real-life, new, and unseen datasets can be calculated. Further, We applied stratified sampling in cross-validation for having a similar distribution of classes in each fold to that in the overall distribution. In our scenario, it is specifically relevant, and that is when working with an unbalanced distribution of classes, and such an unbalance can hinder a model's generalizability. By having a similar distribution in each fold to that of the overall distribution, we make our model evaluation valid.

*Evaluation Metrics*

Performance evaluation of a model is an important part of a machine learning pipeline, and choosing proper evaluation metrics is important in knowing how effectively our model is working. In our analysis, we use a variety of important evaluation metrics, such as Precision, Recall, F1-score, and ROC-AUC.

*Precision* is a proportion of positive prediction out of all positive predictions produced by a model. Precision informs about a proportion of predicted positive cases that occur, and for that reason, it is a significant measure in scenarios when a high cost of a false positive is incurred. For instance, in fraud, high precision will imply that detected fraud wallets will most likely be fraud ones.

*Recall*, in contrast, estimates a proportion of actual positive cases for which a positive prediction is generated. Recall tests a model's performance at predicting all cases of relevance, and its success in terms of positive case capture, in our analysis case in point fraud and suspicious wallets. High recall in our analysis is important in a manner that maximizes fraud and suspicious wallets captured, even at a loss in terms of including a proportion of incorrect ones.

The *F1-score* is a harmonic mean between recall and precision, providing a single value that harmonizes both of them. It is a convenient metric when dealing with imbalanced datasets, providing a fuller picture of model performance concerning both recall and precision taken alone. With a high F1 score, one can observe that a model is selecting relevant cases accurately and rejecting both false positives and negatives effectively.

*Results and Analysis*

*Model Performance*

*Logistic Regression Modelling*

By implementing code snippets in Python that trained and tested a Logistic Regression model with sci-kit-learn. It scaled training and testing feature sets with Standard-Scaler to normalize feature values first. It then constructed a Logistic Regression model with a constant random state for reproduction and trained it with scaled training feature sets. It then created a prediction with scaled testing feature sets. It tested its performance with a range of metrics: a confusion matrix was printed for visualization of true/false positive/negative counts, a classification report was printed for providing precision, recall, F1-score, and support for each label, and overall accuracy was calculated and printed. All these allowed a complete evaluation of the trained Logistic Regression model, checking its performance at a fine-granular (confusion matrix and classification report) and a high level (accuracy).

*Output*

Table 1. Logistic Regression Classification Report

```
Classification Report:
              precision    recall  f1-score   support

           0       0.80      1.00      0.89      2044
           1       1.00      0.00      0.00       503

    accuracy                           0.80      2547
   macro avg       0.90      0.50      0.45      2547
weighted avg       0.84      0.80      0.72      2547

Accuracy: 0.80
```

The above table shows a logistic model evaluation with a confusion matrix and a classification report. In a confusion matrix, 2,044 correct true negatives (class 0) 502 incorrect negatives (class 1), and 1 incorrect positive (false positive) were not detected by the model. In a classification report, a 0.80 precision for class 0 and a perfect 1.00 recall for class 0 mean 80% of predicted instances for class 0 and all actual instances for class 0, respectively, were correct. For a class 1, a much lower 0.00 precision and a 0.50 recall mean that a model is not effective in predicting positive instances and identifies only 50% of them accurately. For class 1, an F1-score of 0.45 is low, and it reflects a performance imbalance between the two classes. In essence, a model accuracy of 0.80 for a whole dataset indicates that a model is effective for a class 0 but its weakness in predicting a class 1 reduces its effectiveness, and it can be seen through a low recall and a low precision for a positive class.

*Random Forest Modelling*

The implemented code snippet trained and evaluated a Random Forest Classifier. It initialized a Random Forest Classifier with 100 trees (n-estimators=100) and a fixed random state for reproducibility. The classifier was trained on the training data (X-train, y-train). Predictions were then made on the test data (X-test). The performance of the trained model was evaluated using a confusion matrix, classification report (including precision, recall, F1-score, and support), and accuracy score. These metrics were printed to the console, providing a comprehensive assessment of the Random Forest model's performance, including insights into its ability to correctly classify different categories and its overall accuracy on unseen data.

*Output*

**Table 2. Random Forest Classification Report**

```
Classification Report:
              precision    recall  f1-score   support

           0       0.80      0.97      0.88      2044
           1       0.13      0.02      0.03       503

    accuracy                           0.78      2547
   macro avg       0.46      0.49      0.45      2547
weighted avg       0.67      0.78      0.71      2547

Accuracy: 0.78
```

The performance of a Random Forest classifier is represented in a table in its classification report and its confusion matrix. In its confusion matrix, it predicted 1,975 actual negatives (class 0) accurately and predicted 60 cases inaccurately as positive, but only detected 10 actual positive cases out of 493 actual positive cases, and 483 cases inaccurately predicted them as not positive. In its classification report, its accuracy for class 1 is 0.13, with 13% of its predicted cases for class 1 have been correct, with a high level of difficulty in predicting positive cases accurately. Recall for class 1 is 0.02, with its model having predicted 2% of actual positive cases accurately. Poor performance is also experienced in its F1-score for class 1, with a value of 0.04, with a high level of imbalance in its performance.

*Support Vector Machines*

The executed code snippet trained and evaluated the Support Vector Machine (SVM) classifier with a radial basis function (RBF) kernel. It initialized an SVM classifier (SVC) with the specified kernel and a random state for reproducibility. The classifier was trained using the training data (X-train, y-train). Predictions were then made on the test data (X-test). The performance of the trained SVM model was evaluated using a confusion matrix, a classification report (containing precision, recall, F1-score, and support for each class), and the accuracy score. These evaluation metrics were printed to the console, providing a comprehensive view of the SVM model's performance, including its classification accuracy and its ability to distinguish between different classes.

*Output*

Table 3. Support Vector Machines Classification Report

```
Classification Report:
              precision    recall  f1-score   support

           0       0.80      1.00      0.89      2044
           1       0.00      0.00      0.00       503

    accuracy                           0.80      2547
   macro avg       0.40      0.50      0.45      2547
weighted avg       0.64      0.80      0.71      2547

Accuracy: 0.80
```

The above table presents a performance evaluation of a Support Vector Machine (SVM) model in terms of a classification report and a confusion matrix. According to a confusion matrix, 2,044 actual negatives (class 0) and no positive cases (class 1) were detected accurately, and 503 actual positive cases and no actual positive cases were detected in terms of actual positive cases and false positive cases, respectively. According to a classification report, a 0.80 for an 80% recall for class 0, signifies 80% of instances predicted for class 0, and a perfect 1.00 recall for class 0, meaning all actual instances for class 0 were detected accurately. Conversely, a 0.00 for a 0.00 recall for class 1, means no actual positive cases detected, and an F1-score of 0.00 for a 1 for class 1, meaning no actual positive cases detected, with a model accuracy of 0.80, but a somewhat wrong figure, in that it reflects a model with high accuracy in terms of detecting predominantly a model capable of detecting a class 0 but with high failure in terms of positive case detection in future implementations, and therefore, an improvement in terms of positive case capture in future implementations is warranted.

*Comparison of All Models*

The computed code script performed a comparative analysis of a variety of machine learning algorithms (Random Forest, Logistic Regression, and SVM) and identified the best model in terms of its performance in terms of its best F1-score. It formed a function evaluation model that calculated and returned a dictionary of evaluation statistics including accuracy, precision, recall, and F1-score, concerning actual and predicted labels. It then iterated over trained models, storing evaluation values in a list with title results. The list was then converted to a Pandas Data-Frame comparison_df for easier comparison. Data Frame was sorted in descending order concerning its best F1-score for model ranking, and comparison values were printed out. Finally, it identifies and prints out the best model in terms of its best F1 score, providing a simple summary of the best performance model out of the ones evaluated.

*Output*

```
Model Comparison Results:
                 Model  Accuracy  Precision    Recall  F1 Score
1        Random Forest  0.779348   0.126582  0.019881  0.034364
0  Logistic Regression  0.802905   1.000000  0.001988  0.003968
2                  SVM  0.802513   0.000000  0.000000  0.000000
```

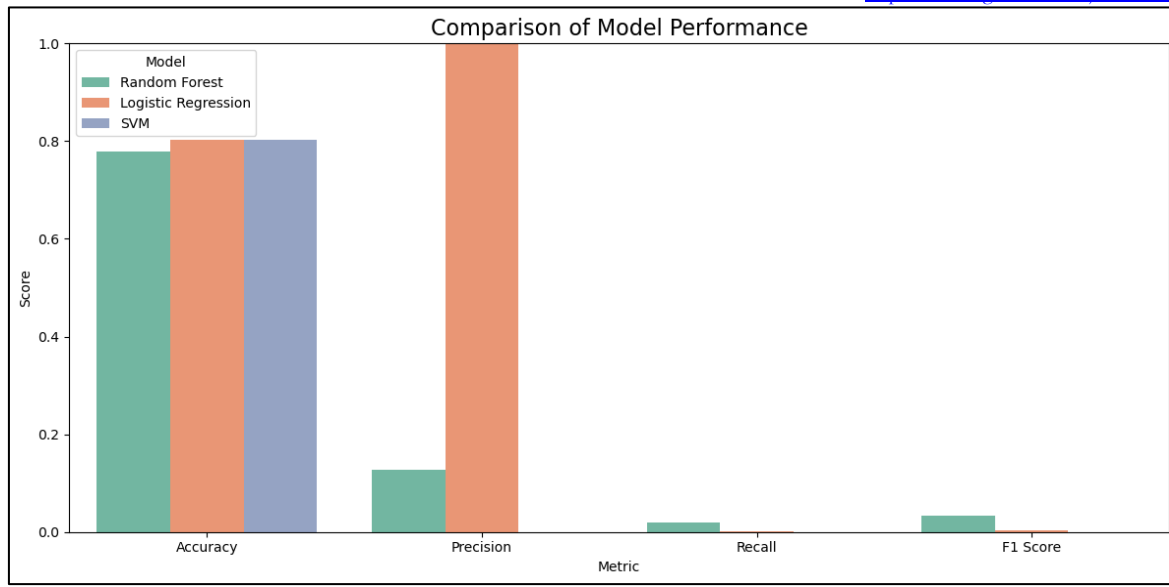

**Figure 7. Comparison of Model Performance**

The above graph is a comparison of three models, namely, Logistic Regression, Random Forest, and Support Vector Machine (SVM), in terms of performance in terms of accuracy, recall, precision, and F1 score. With an accuracy of 77.99%, a recall of 0.019881, and a precision of 0.126582, the Random Forest model achieved an F1 score of 0.034364, poor effectiveness in predicting positive cases in terms of overall accuracy, even when it performed with high overall accuracy. Logistic Regression, on its part, performed poorly with an accuracy of 0.003968, with its failure to predict any positive cases, with its recall, precision, and F1 score at 0.003968, respectively. Similarly, even SVM performed beyond average with an accuracy of 80.21%, but with no positive cases predicted, and its recall, precision, and F1 score at 0.000000, respectively. What can be noticed is that even with high overall accuracy, with its best F1 score amongst the three, for its case, its performance in effectively predicting positive cases is poor, and hence, even with high overall accuracy, its effectiveness in predicting positive cases is poor, with a depiction of difficulty in predicting positive cases in such a classification problem. Random Forest emerged as the best model with the highest F1 Score, showcasing its ability to handle non-linear relationships in the data. Insights revealed significant patterns in wallet activity, such as the correlation between unredeemed transactions and final balances.

*Insights on Suspicious Transactions*

Analysis of suspicious activity in cryptocurrency networks reveals a variety of key trends and behaviors that can serve as markers for suspicious activity. High volumes of activity in a single wallet over a relatively short period is one such key trend seen, and such activity tends to represent a "smurfing" money laundering scheme in which illicit funds are broken down into small values and moved through a sequence of wallets in an attempt to make them untraceable in origin. High volumes of activity, when seen in clusters and then not seen for a period, demand a high level of investigation. Trends in values in transactions can also reveal important information; for instance, wallets consistently transacting similar values can represent an attempt at creating a cover for legitimacy, particularly when such values closely coincide with predefined values that will draw regulators' scrutiny. In addition, analysis of trends in activity reveals suspicious behavior such as mixing service use—funds aggregated and then redistributed, and becoming untraceable in origin— frequently in conjunction with wallets with a record of high-risk activity in the past, or with ties to known illicit groups. By combining these behaviors, an overall picture can be constructed of suspicious wallets, and potential threats can be actively discovered and addressed early, even at a relatively early stage.

The analysis then continues with an examination of high-risk wallets and transaction behavior and with an examination of connectivity about suspicious activity in a network. High-risk wallets, for instance, have a record of transacting with other entities that have been flagged for suspicious activity, and mapping out such connectivity helps visualize relationships in a network. For instance, a wallet with a record of transacting with a range of addresses with a proven record for activity in dark web marketplace activity, such as ransomware attack activity, raises an alarm. Scrutiny of such activity can then unveil additional suspicious activity such as timing, location, and goods and services transacted. Wallets with a high level of activity with new addresses, a common fraudster practice in an attempt to go undetected, represent a high level of suspicious activity. Examining clusters of such high-risk activity, it is then feasible to map out larger networks of illicit activity and allow for timely intervention in suspected threats. Information gained through such analysis proves incredibly useful, not only in its role in suspicious activity identification, but in providing a deeper view of activity in and about a cryptocurrency ecosystem, and in enhancing security and integrity in a network overall.

*Use Cases*

Practical examples of such fraud detected serve to illustrate the utility of such behavior and patterns in fraud and risk detection and avoidance in terms of cryptocurrency transactions. One such use case involved a high level of transactions with many new addresses in a short period for a specific wallet detected. On deeper analysis, such transactions were determined to form part of a larger scheme for money laundering, with such funds derived through a phishing attack being channeled through such a scheme. Such a wallet's transaction behavior included repetitive, small transfers, below such thresholds in traditional banking systems, in most cases, such behavior generated an alarm. Such a case is an example of fraudsters exploiting the fact that cryptocurrencies lack a single, monolithic controlling entity, such that such behavior could not be monitored, with techniques such as layering being used to make such funds untraceable in origin.

A similar case in point is tracking wallets involved in ransomware activity. In such an instance, such activity between suspected ransomware wallets was analyzed, and specific behavior of rapid transactions with a variety of addresses that appeared to have dealings with real companies in such activity emerged. The analysis determined such companies processed payments for illicit activity unaware, in a sense, acting almost in an intermediary role for such activity. High-risk wallets' activity in sending and receiving payments with mixing services added a level of challenge for tracking activity. By creating such a nexus, such companies could then act and notify, and such additional illicit activity could be circumvented, and such additional exposure to such fraud could potentially be circumvented. Not only do such use cases demonstrate effectiveness in using such sophisticated analysis techniques to expose such fraud behavior, but such use cases demonstrate a larger impact for such use in terms of compliance with regulators and a necessity for additional monitoring tools in such a changing environment for cryptocurrencies.

*Practical Applications*

*Implications for financial security*

The application of machine learning algorithms for tracking Bitcoin transactions carries significant implications for financial security in a cryptocurrency marketplace. To effectively use such technology, financial institutions, and exchanges will have to implement complex predictive algorithms with real-time analysis capabilities for tracking suspicious activity. Historical transactional information can be trained in such algorithms to identify fraud patterns for such, including anomalous volumes, rapid multi-transfer activity, and behavior not in compliance with a documented behavior model for a specific individual. By employing techniques such as clustering and anomalous behavior, financial institutions can tag for investigation transactions with a high level of suspicious activity, allowing for proactive fraud intervention. Including a feedback mechanism in which a human can review and update the output of a model will enable increased accuracy over some time, dynamically updating for new emerging threats in a changing environment.

The intelligence derived through suspicious transaction analysis can also become useful guidance for financial companies and cryptocurrency exchanges in fraud preventive process improvement. By creating rich profiles of users through behavior, transaction behavior, and risk analysis, companies can implement multi-leveled scrutiny for certain segments of users. For example, high-value users, including high-value transactors with suspicious wallets and behavior for money laundering, can have additional processes for due diligence, including identity checking and caps in transactions. Besides, creating collaboration with regulators and law enforcement can allow for information sharing regarding emerging trends and new types of cryptocurrency threats. By creating an effective system through the integration of machine analysis with strong processes for collaboration and user verification, financial companies can effectively counter fraud and make overall security in cryptocurrencies much more powerful.

*Integration with Regulatory Frameworks*

The information derived through Bitcoin transaction analysis can go a long distance in assisting U.S. regulators in curbing financial crimes committed with cryptocurrencies. By leveraging machine learning algorithms and combining them with regulatory frameworks, regulators can develop smarter compliance tools that can monitor suspicious activity in real time, both at individual platforms and between platforms. Not only can such a mechanism simplify fraud detection, but it can also simplify acting in a proactive compliance manner, too. By leveraging information derived through analysis of transaction behavior, regulators can utilize them to inform guidelines and best practice development for cryptocurrency platforms, such that such platforms have robust anti-money laundering (AML) and know-your-customer (KYC) processes in place. By collaborating with machine learning professionals, regulators can make analysis capabilities smarter, such that regulators can outwit sophisticated fraudsters' methodologies and update laws to counter new and emerging threats.

To report and monitor in real-time, regulators can make use of a variety of tools and techniques optimized for ease of compliance processes. For one, the use of reporting tools with triggers for suspicious activity reporting, following predefined thresholds, can mitigate the work involved in tracking activity through a manual burden. With added integration with sophisticated analysis, such tools can assess factors for risk and generate in-depth reports for regulators' review. With partnerships with blockchain analysis companies, regulators can access powerful tools for mapping out transaction flows and identifying relationships between entities, strengthening investigation capabilities. By combining such technology with operational frameworks, regulators can develop a strengthened mechanism for supervision, with transparent and secure operations in cryptocurrency markets and effective financial crime fighting.

*Scalability to Other Cryptocurrencies*

The observations and techniques derived through Bitcoin transaction analysis can then be translated to Ethereum and Litecoin and then to a broader range of assets applicable for use. All cryptocurrencies have idiosyncratic behavior and individualistic traits in transactions, but most of the deep-rooted fraud patterns, such as money laundering and phishing, will have similar profiles no matter the platform. By mapping over algorithms for specific idiosyncratic behavior in each cryptocurrency, financial regulators and financial institutions can have a broader range of assets under one fraud system with a high level of adaptability for idiosyncratic behavior and fraud profiles specific to individual cryptocurrencies. For instance, Ethereum's smart contract feature can have new fraud avenues, and therefore, individual models will have to monitor activity in and out of dApps and between wallets.

Retrospectively, such an approach is facilitated with growing access to tools for cross-chain analysis of data, and one can monitor transactions between a range of networks of blockchains with such tools. With such tools, one can utilize them to monitor relations between cryptocurrencies, and thus, expose trends not seen when one considers individual assets in a vacuum. By having a big-picture view of cryptocurrency transactions, one can have a deeper view of the overall financial environment and detect vulnerabilities extending over a range of cryptocurrencies. Not only will such an integrated view secure overall fraud deterrent, but will instill growing confidence in both investors and users in the cryptocurrency marketplace and, and the long run, contribute towards its growth and solidity.

## Discussion and Future Directions

*Impact on Cryptocurrency Monitoring in the USA*

The application of machine algorithms in tracking cryptocurrencies is a tool for creating transparent and secure U.S. markets. As virtual currencies gain increased acceptance and transactions become increasingly sophisticated, machine algorithms can provide processing capabilities for enhancing supervision and compliance operations. Complicated algorithms can be programmed to search through massive sets of transactional information, identifying trends that could be indicative of fraud and compliance failures. With the use of past data, such algorithms can become trained to detect abnormalities in real time, and regulators and financial institutions can respond promptly to suspicious activity. Not only will such a proactive action go towards curbing financial fraud, but will also improve overall integrity in virtual currency environments, and in return, stimulate increased acceptance and use.

The worth of AI-powered fraud detection extends even deeper, with added positive ramifications for financial security in the cryptocurrency marketplace. As algorithms for machine learning become ever more capable of distinguishing between real and suspicious transactions, cryptocurrency platforms can make operations even safer. For one, through identifying and countering money laundering and manipulation-related dangers, such platforms can protect themselves from financial penalties and loss of goodwill. In addition, a safer environment inspires confidence in investors, welcoming institution investors who, previously, have avoided taking part in the cryptocurrency markets for concern about fraud and price manipulation. Overall, the use of machine learning in tracking cryptocurrencies is an important step toward a safer, transparent, and secure financial environment, with a positive impact on everyone involved in the ecosystem.

*Limitation and Challenges*

Although a lot of potential for enhancing the tracking of cryptocurrencies exists in machine learning, a variety of key weaknesses and obstacles must first be addressed for its efficacy to dominate. Perhaps its most significant issue concerns data quality. Well-formed and complete datasets must exist for model training, but cryptocurrencies' decentralized state tends to fragment datasets into disconnected sources. Incomplete and uncorrelated transactional information, reporting gaps, and reporting standards discrepancies can cause model prediction inaccuracies. Complications regarding safeguarding individual information and its integrity, such as in sensitive user information, represent a critical issue, and a balancing act between effective tracking and safeguarding individual freedom must take place, with robust frameworks in place for compliance with information safeguard laws.

Another challenge is the interpretability of machine algorithms. As strong, as such algorithms can generate, their complexity tends to make them "black boxes," and it can become difficult for stakeholders to understand decision processes. Transparency can suffer, and trust in technology, in specific, with regulators and financial entities that require transparent justification for actions taken concerning model output, can suffer in consequence. Besides, with continuous development in fraud methodologies, refreshing machine algorithms to detect new methodologies is a continuous challenge. Hackers become smarter, and methodologies can evade present tools with ease. Repeated retraining and model refresh become imperative, but expensive, and demand concentrated efforts to have a head start over new threats emerging.

*Opportunity for Future Studies*

The field of tracking cryptocurrencies holds a lot of room for future investigation, particularly in studying the union of blockchain analysis and machine learning techniques. By combining both techniques, one can develop stronger frameworks with a heightened fraud detection capacity. With its level of detail in tracking activity flows, blockchain analysis can target suspicious activity through observable trends in activity. By combining them with machine learning algorithms, one can develop an overall toolset not only for fraud detection but also for attributing suspicious activity to a specific actor in a blockchain environment. With such a union, one can have a fairer chance at effective compliance and fraud protection, and, subsequently, a safer marketplace in general.

Moreover, a high opportunity is present for developing hybrid methodologies blending both statistical methodologies and machine learning algorithms for real-time and high accuracy regarding anomalous behavior detection. Hybrid methodologies can exploit strengths in methodologies such as utilizing rule-based methodologies for first-level filtering and machine learning for in-depth analysis of suspicious transactions. Hybrid methodologies' adaptability can make real-time, dynamic adaptations towards changing fraud techniques, and therefore, allow for sustained effectiveness in a changing environment. Federated methodologies for training machine learning algorithms in a decentralized network can be designed in future work, and training can be performed in a manner that keeps individual user information private. By developing these, the field can make considerable improvements, and a new level of effectiveness can be attained in terms of security and integrity in cryptocurrency transactions.

## Conclusion

The prime objective of this research project is to develop machine learning algorithms capable of effectively identifying and tracking suspicious activity in Bitcoin wallet transactions. With high-tech analysis, the study aims to create a model with a feature for identifying trends and outliers that can expose illicit activity. The current study specifically focuses on Bitcoin transaction information in America, with a strong emphasis placed on the importance of knowing about the immediate environment in and through which such transactions pass through. The dataset is composed of in-depth Bitcoin wallet transactional information, including important factors such as transaction values, timestamps, network flows, and addresses for wallets. All entries in the dataset expose information about financial transactions between wallets, including received and sent transactions, and such information is significant for analysis and trends that can represent suspicious activity. This study deployed three accredited algorithms, most notably, Logistic Regression, Random Forest, and Support Vector Machines. In retrospect, Random Forest emerged as the best model with the highest F1 Score, showcasing its ability to handle non-linear relationships in the data. Insights revealed significant patterns in wallet activity, such as the correlation between unredeemed transactions and final balances. The application of machine algorithms in tracking cryptocurrencies is a tool for creating transparent and secure U.S. markets. As virtual currencies gain increased acceptance and transactions become increasingly sophisticated, machine algorithms can provide processing capabilities for enhancing supervision and compliance operations. Complicated algorithms can be programmed to search through massive sets of transactional information, identifying trends that could be indicative of fraud and compliance failures. With the use of past data, such algorithms can become trained to detect abnormalities in real-time, and regulators and financial institutions can respond promptly to suspicious activity.